\documentclass[a4paper, 10pt, conference]{ieeeconf}      
\usepackage{FG2026}
\usepackage{gensymb}
\FGfinalcopy 

\IEEEoverridecommandlockouts                              
\usepackage{graphics} 
\usepackage{epsfig} 
\usepackage{mathptmx} 
\usepackage{times} 
\usepackage{amsmath} 
\usepackage{amssymb}  
\usepackage{lipsum} 
\usepackage{multirow}
\usepackage{makecell}
\usepackage{booktabs}
\usepackage{paralist}
\usepackage{multirow} 
\usepackage{fancyhdr}

\usepackage{booktabs}
\def\FGPaperID{436} 

\title{\LARGE \bf
Event-based Liveness Detection using Temporal Ocular Dynamics: ~ ~ An Exploratory Approach
}

\author{\parbox{16cm}{\centering
    {\large Nicolas Mastropasqua$^{1,2}$ \quad
    Ignacio Bugueno-Cordova$^{3,4}$ \quad
    Rodrigo Verschae$^{5}$ \quad \\
    Daniel Acevedo$^{1,2}$ \quad
    Pablo Negri$^{1,2}$ \quad 
}\\
    {\normalsize
    $^1$ Universidad de Buenos Aires, Facultad de Ciencias Exactas y Naturales, Dpto. de Computación, Argentina \\
    $^2$ CONICET-UBA, Instituto de Ciencias de la Computación (ICC), Argentina \\
    $^3$ Department of Electrical Engineering, Universidad de Chile, Chile \\
    $^4$Institute of Engineering Sciences, Universidad de O'Higgins, Chile\\
    $^5$Department of Informatics, Universidad Técnica Federico Santa María, Chile\\
}}
    \thanks{This work was partially supported by project UBACyT 20020220200147BA from Universidad de Buenos Aires and the FONDEQUIP project EQM17004. 
}}

\begin{document}

\ifFGfinal
\thispagestyle{empty}
\pagestyle{empty}
\else
\author{Anonymous FG2026 submission\\ Paper ID \FGPaperID \\}
\pagestyle{plain}
\fi
\maketitle

\thispagestyle{fancy}
\renewcommand{\headrulewidth}{0pt}
\fancyhf{}


\begin{abstract}
Face liveness detection has been extensively studied using RGB cameras, achieving strong performance under controlled conditions but often failing to generalize across sensors and attack scenarios. In this work, we explore event cameras as an alternative sensing modality for liveness detection based on temporal ocular dynamics.
Event cameras capture sparse, asynchronous changes in brightness with microsecond resolution, enabling precise analysis of fast eye movements such as saccades. Replay attacks cannot faithfully reproduce these dynamics due to temporal resampling and display artifacts, leading to distinctive spatio-temporal patterns in the event domain.
We design a data collection protocol to extend RGBE-Gaze with replay-attack recordings, yielding an event-based fake counterpart for liveness detection.
We analyze event-driven temporal features from eye regions and evaluate their effectiveness for ocular motion segmentation and liveness classification.
Our results show that event-based representations enable reliable discrimination between genuine and replayed sequences, achieving up to 95.37\% top-1 accuracy with a spiking convolutional neural network. These preliminary findings highlight the potential of event-based sensing for robust and low-latency liveness detection.
\end{abstract}

\section{Introduction}

Face liveness detection has been extensively studied under the RGB imaging modality~\cite{apgar2021,khairnar2023,yu2023}. A large body of work using conventional RGB cameras reports excellent results in intra-dataset evaluation protocols under controlled settings~\cite{sharma2023survey,zhang2020adversarial,alshaikhli2021,George2019DeepPB,Lucena2017TransferLU}. However, these methods frequently fail to generalize, as they tend to rely on dataset-specific artifacts, such as illumination conditions, presentation attack instruments, or sensor characteristics, rather than intrinsic liveness cues. Consequently, performance drops significantly in cross-dataset evaluations on unseen attack scenarios. This limitation motivates the exploration of sensing modalities that capture more robust and intrinsic characteristics of genuine human motion.

\begin{figure}[t]
    \centering
    \includegraphics[width=\linewidth]{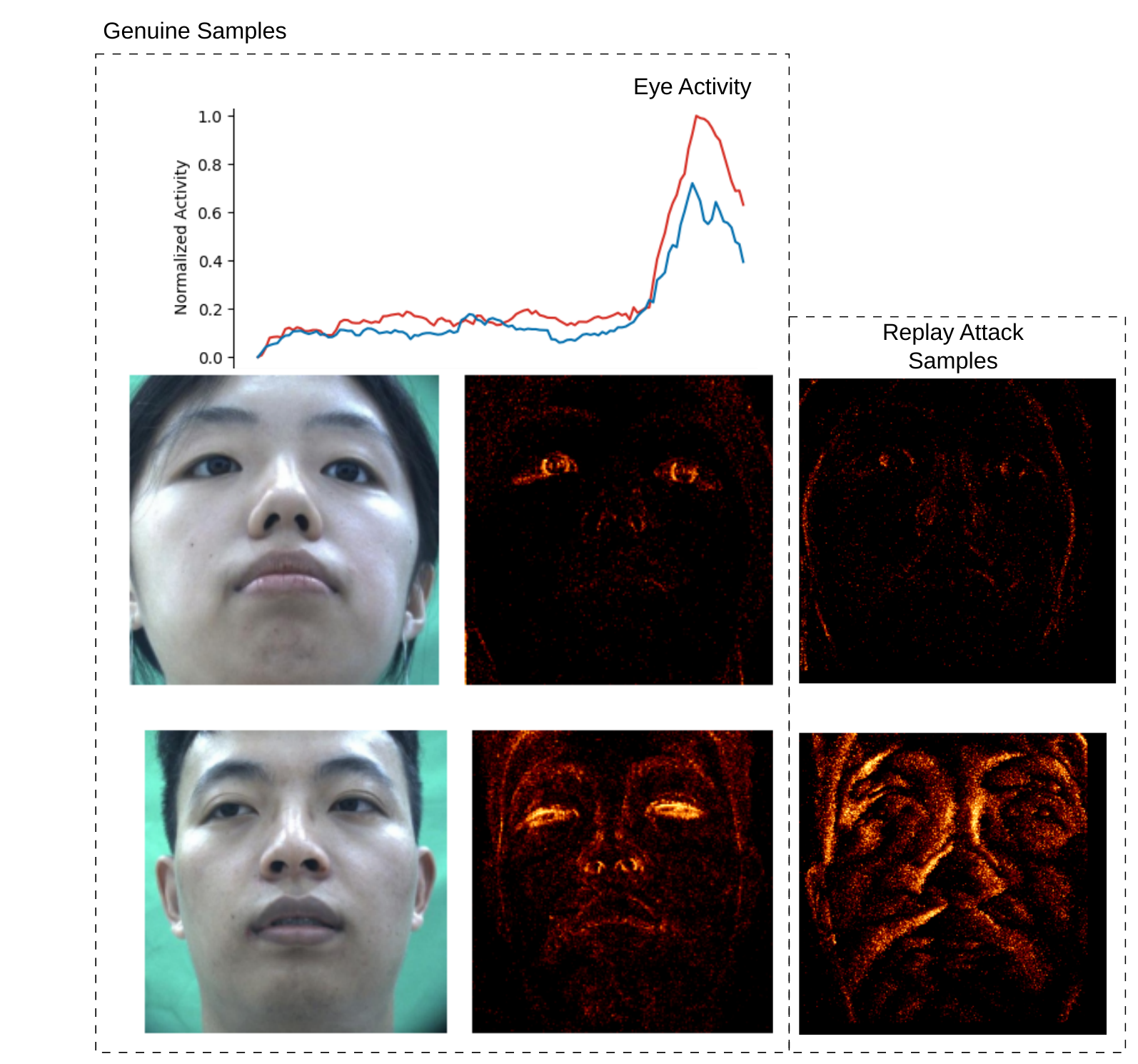}
    \caption{
Samples from the RGBE-Gaze dataset extended with our replay attack recordings. Each row shows a genuine RGB frame, its corresponding Surface of Active Events (SAE) over a 200 ms window, and the SAE of a replay attack. The top plot shows the eye activity for the subject in the first row over the same temporal window}
    \label{fig:dataset_samples}
\end{figure}

Several approaches have explored complementary modalities, including depth~\cite{8272713}, near-infrared~\cite{8575245}, and thermal imaging~\cite{Sun2011TIRVISCF}. In this context, event cameras offer a fundamentally different sensing paradigm for addressing the problem, capturing asynchronous brightness changes with microsecond resolution, low latency, and high dynamic range~\cite{Mead1993ASM, Gallego_2022}. These properties make event cameras particularly well-suited for capturing fast motion patterns while preserving user privacy.

\begin{figure*}[!h]
    \centering
    \includegraphics[width=\linewidth]{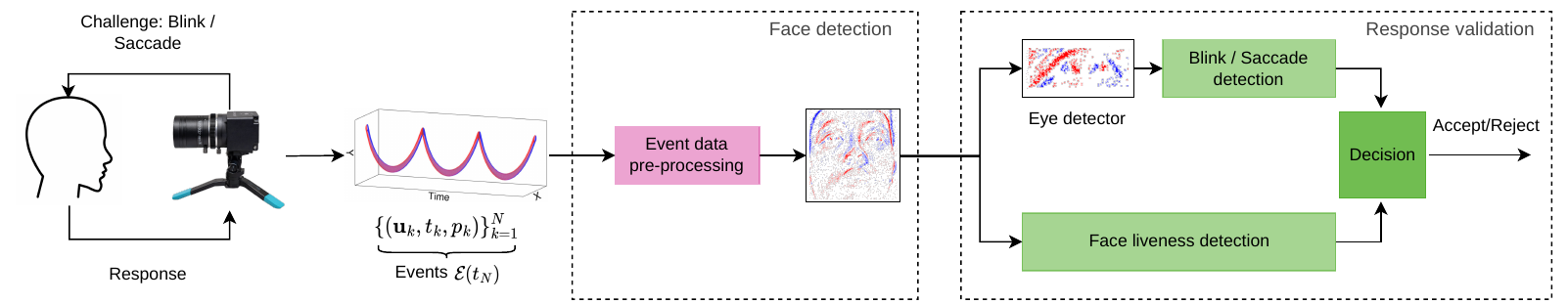}
    \caption{Proposed pipeline for event-based liveness detection using temporal ocular dynamics. A challenge–response mechanism enforces real-time responses to improve robustness against replay and synthetic attacks. Event streams are preprocessed to extract face and eye regions using synchronized RGB frames. Temporal features detect blinks or saccades, while a parallel branch performs face liveness detection. The outputs are combined to produce the final decision.}
    \label{fig:pipeline}
\end{figure*}

Ocular movements offer a promising source of subtle liveness cues. Fast eye motions, such as saccades and microsaccades, occur at millisecond timescales and are difficult to reproduce accurately in replay attacks due to temporal resampling and display artifacts. While prior work~\cite{iddrisu2024eventeyesurvey} has explored event-based eye analysis in near-eye scenarios, primarily for eye tracking and gaze estimation, their application for liveness detection in remote-eye settings remains largely unexplored. 

In this work, we propose an active event-based anti-spoofing approach targeting replay attacks. The pipeline begins with an initial challenge–response phase, which prompts the user to execute a rapid ocular movement. Upon successful detection of face and eye regions, the system performs simultaneous ocular movement and face liveness detection by analysing distinctive event-based temporal patterns.

The main contributions of this work are:
\begin{itemize}
\item A novel application of event cameras as a sensing modality for face liveness detection. To the best of the authors’ knowledge, this is the first study to investigate event-based liveness detection leveraging temporal ocular dynamics in remote-eye scenarios.
\item An event-based method for the temporal segmentation of saccades and blinks in remote-eye settings.
\item A replay attack dataset, derived from RGBE-Gaze, designed to advance research in event-based liveness detection.
\end{itemize}

The extended dataset and codes will be made available on the project website\footnote{https://ev-latam.github.io/event-face-liveness}.


\section{Related Work}

\subsection{Event-based Face Analysis and Liveness Detection}

Event-based facial analysis has gained increasing attention as an alternative paradigm for capturing dynamic visual information  \cite{becattini2025neuromorphicfaceanalysissurvey,iddrisu2024eventeyesurvey}. Prior work has explored face-related tasks using event cameras, including detection, eye blinking, and facial expression recognition.

Face detection in this domain has been addressed using both temporal and learning-based approaches. Early work in \cite{barua2016} introduced patch-based models, while \cite{lenz2020} leveraged blink dynamics for probabilistic face localization and tracking. Learning-based methods include kernelized correlation filters \cite{ramesh2020} and datasets such as FES \cite{bissarinova2024fes}, enabling real-time detection. Multispectral approaches further improve robustness compared to conventional imaging \cite{himmi2024msevs}.

Facial expression recognition has also been explored using neuromorphic representations and methods \cite{barchid2024spikingfer, berlincioni2023nefer, verschae2023eventgesturefer,mastropasqua2025iccvw,mastropasqua2025icprs,verschae2026evtransfer}. In this context, Spiking Neural Networks (SNNs) achieve competitive performance while improving efficiency, showing that such data captures subtle facial dynamics not easily observed with frame-based sensors.

Despite these advances, the application of event cameras to liveness detection remains limited. The most closely related work is NeuroBiometric \cite{Chen2021NeuroBiometricAE}, which proposes an authentication system based on eye-blink dynamics with 94.8\% accuracy. However, it is restricted to near-eye configurations and—critically—does not address presentation attacks, leaving the role of event-based ocular dynamics in remote-eye liveness detection largely unexplored. To the best of our knowledge, this is the first work to leverage temporal ocular dynamics from event data for replay attack detection in remote-eye scenarios characterized by lower spatial resolution, head motion variability, and unconstrained environments.

\subsection{Event-based Ocular Movement Analysis}
In the past few years, the study of fast eye movements with event cameras has gained increasing attention due to their outstanding temporal resolution and low power consumption, enabling potential applications in AR/VR headsets and biomedical settings, among others. Foundational work by Angelopoulos et al.~\cite{angelopoulos2021} demonstrated that it was possible to estimate gaze vectors at an equivalent rate of 10 kHz using a hybrid method with events and frames in a near-eye setup.

The release of large publicly available near-eye datasets, such as 3ET~\cite{3et} and EV-Eye~\cite{ev-eye}, has enabled the comprehensive study of eye-tracking-related tasks. Recent works have focused on high-frequency gaze estimation and tracking using hybrid and fully event-driven models \cite{wu2025brat, huang2025exploring, truong2025dual,10.1145/3649902.3653471}, leveraging the high temporal resolution of event sensors to capture rapid eye movements. Among these approaches, studies such as \cite{Groenen2025GazeSCRNNEN,10678580} have shown that Spiking Neural Networks (SNNs) can effectively learn the spatio-temporal dynamics of fast eye movements in this context. Furthermore, some works have explored eye-blink dynamics for face detection and face authentication applications~\cite{Chen2021NeuroBiometricAE,lenz2020}.

However, the vast majority of existing works overlook the problem of saccade detection and precise temporal segmentation. Voluntary saccades are ballistic eye movements that rapidly shift gaze from one target to another, reaching speeds of approximately 300~$\degree/\text{s}$. Their short duration makes them well suited for event-based cameras, as saccades with amplitudes less than 18$\degree$ typically last only 20--80~ms~\cite{Baloh1975QuantitativeMO}. Previous work~\cite{Iddrisu2026EyeMC} has addressed this problem using manually annotated saccades and fixation periods in the EV-Eye dataset, but focused primarily on detection rather than precise temporal segmentation.

Despite these efforts, the analysis of rapid eye movements in remote-eye scenarios—where the camera is positioned at least 50 cm from the eyes—remains largely unexplored. To date, RGBE-Gaze~\cite{rgbe} appears to be the only comprehensive dataset designed for remote-eye saccade analysis. Remote-eye settings constitute a particularly relevant operating condition for real-world applications, as they enable unconstrained head movement and contactless interaction in domains such as biometrics, affective computing, and driver monitoring systems. However, these scenarios introduce additional challenges compared to near-eye configurations, including low-resolution eye regions, head pose variability, and changing illumination conditions.


\section{Experimental Setup \& Dataset Collection}

We propose an active liveness system based on event cameras, summarized in Figure~\ref{fig:pipeline}. In the first phase, the user is prompted to perform a fast ocular movement, such as a saccade or blink. This challenge–response mechanism is designed to improve robustness against replay and synthetic attacks by enforcing real-time responses. Upon successful detection of the face and eye regions, the system addresses two tasks concurrently:
\begin{inparaenum}[a)]
\item detecting the occurrence of the eye movement, and
\item determining whether the input corresponds to a genuine or replayed sequence.
\end{inparaenum}

We assume that an attacker attempting to replay event data faces inherent limitations, as the attack would require reconstructing the event stream into a frame-based representation for display, which introduces temporal and spatial distortions. Instead, we consider a stronger attack scenario in which the adversary has access to RGB recordings aligned with the original event data. Based on this assumption, we define a data collection protocol to evaluate the system and introduce a replay attack dataset derived from RGBE-Gaze~\cite{rgbe}. 

The face detection phase is simplified by extracting facial and eye bounding boxes using dlib from the aligned genuine RGB stream. For the event replay clips, we apply an affine transformation to map the extracted landmark coordinates to screen coordinates. Although more sophisticated event-based alternatives exist, we restrict our analysis to this setup for the sake of simplicity.

\subsection{Genuine dataset: RGBE-Gaze}
The RGBE-Gaze dataset by Zhao et al.~\cite{rgbe} is a multimodal dataset designed for remote gaze tracking using spatio-temporal synchronized RGB (FLIR BFS-U3-16S2C) and event (Prophesee EVK4) cameras, totalling 66 subjects (37 male and 29 female) aged between 18 and 28 years. The experiments were conducted with the subject's head free of any constraint, sitting in front of a screen at a distance between 60 and 80 cm. The subjects underwent 6 experimental sessions; In sessions 1-4 (Fixation + Saccade), visual stimulus triggering saccades appeared anywhere on the screen in a random sequence, allowing for fixation periods in between. In sessions 5-6 (Smooth Pursuit), participants were asked to fixate on a stimulus moving across the screen in a predictable, square-wave trajectory. 

\subsection{Video replay Attack Dataset Collection}

\subsubsection{Protocol and Set-up}

The replay video attack recordings were set up in a room with controlled artificial illumination. We used an IPS screen with a resolution of  1920 x 1080 and a refresh rate of 60 Hz as the interface to reproduce the genuine RGB clips. Each clip was centred on the screen, and video playback was performed at the original frame rate and resolution with no rescaling. The background of the screen was set to black, minimizing possible illumination leakage coming from the sides, and the brightness was as low as possible. The screen was placed in front of the cameras at a fixed distance, ensuring the frames occupy roughly the same area of pixels in our event camera as the genuine recordings.

Genuine clips from all subjects were played sequentially. Since the abrupt change between the last and the first frame of contiguous sequences triggers lots of artificial events that would leak into the next sequence, the first frame is paused for 1 second, smoothing the transition. 
We added a pause label far from the ROI of each starting frame to serve as a visual marker for post-temporal segmentation per clip. The first hundred milliseconds of playback after the first frame were dropped to prevent a ``fade-out'' transition from leaking events into the new clip. 

\begin{figure}[!h]
    \centering
    \includegraphics[width=0.9\linewidth]{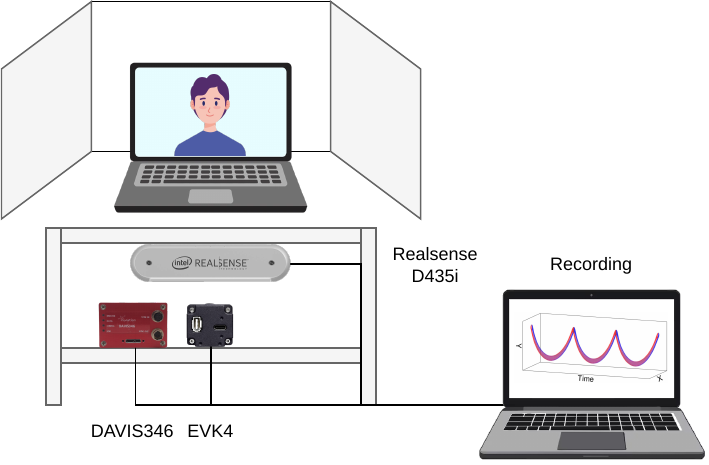}
    \caption{Experimental setup for dataset collection. A display is used to show replayed RGB sequences, while a multi-sensor rig composed of a RealSense D435i, DAVIS346, and Prophesee EVK4 captures RGB and event data.}
    \label{fig:dataset_setup}
\end{figure}

\subsubsection{Sensors}

To acquire this dataset, we used three vision sensors: a RGB-D sensor (RealSense D435i), an Active Pixel Sensor-based event camera (DAVIS346~\cite{brandli2014, mueggler2014}), and an event-based camera (Prophesee EVK4). Table~\ref{tab:vision-sensors-summary} summarizes each sensor's data modality, field-of-view, and spatial resolution.

\begin{table}[h]
    \caption{Relevant technical characteristics of employed vision sensors.}
    \label{tab:vision-sensors-summary}
    \centering
    \renewcommand{\arraystretch}{1.2}
    \begin{tabular}{lllc}
        \hline
        \textbf{Camera model} & \textbf{Data} & \textbf{Field-of-view} & \textbf{Resolution} \\ \hline

        RealSense D435i
        & \makecell[l]{RGB Frame \\ Depth Frame}
        & \makecell[l]{H: 69.0° \\ V: 42.0° \\ D: 77.0°}
        & \makecell[c]{1280 $\times$ 720} \\ \hline

        DAVIS346
        & \makecell[l]{Gray frame \\ Events}
        & \makecell[l]{H: 29.9--113° \\ V: 22.7--99.7° \\ D: 36.9--215°}
        & \makecell[c]{346 $\times$ 260} \\ \hline

        Prophesee EVK4
        & Events
        & \makecell[l]{H: 41.4° \\ V: 23.6° \\ D: 47.0°}
        & 1280 $\times$ 720 \\ \hline
    \end{tabular}
\end{table}

A custom-made capture platform (Figure~\ref{fig:dataset_setup}) was employed to mount the cameras, accounting for their physical dimensions and respective fields of view, and they were controlled via a ROS script to ensure temporal synchronization.

Before data acquisition, all sensors were calibrated to obtain their intrinsic parameters and ensure geometric consistency across modalities. Calibration was performed using standard chequerboard-based procedures, allowing the estimation of focal lengths, principal points, and distortion coefficients for each camera. These parameters were used to align the different sensing modalities and ensure accurate region-of-interest extraction across RGB and event streams. 
The resulting intrinsic parameters are summarized in Table~\ref{tab:calibration-parameters}.

\begin{table}[h]
    \centering
    \small
    \caption{Intrinsic parameters of the employed vision sensors.}
    \label{tab:calibration-parameters}
    \begin{tabular}{lcccc}
    \toprule
    Sensor & $f_x$ & $f_y$ & $c_x$ & $c_y$ \\
    \midrule
    RealSense D435i & 1384.72 & 1385.10 & 960.35 & 540.18 \\
    DAVIS346 & 319.49 & 319.13 & 179.10 & 119.75 \\
    Prophesee EVK4 & 7230.23 & 7230.23 & 627.02 & 802.98 \\
    \bottomrule 
    \end{tabular}
\end{table}

\subsubsection{Dataset description}

We selected a smaller subset of subjects from the genuine dataset for our experimental recordings, as summarized in Table~\ref{tab:dataset_1}. Concretely, we kept a total of 17 subjects (4 female, 13 male) and only the clips corresponding to the first saccade experiment and the last smooth pursuit experiment. To keep the total running time short, since each original saccade experiment contains 60 stimuli and is about 3.5 minutes long, we decided to trim their duration to a third, ensuring at least 17 stimuli were present on average. A few samples comparing genuine and replay attack recordings are shown in Figure~\ref{fig:dataset_samples}, and a summary of the resulting clips in the final dataset is provided in Table~\ref{tab:dataset_2}.

\begin{table}[h]
\centering
\caption{Summary of the proposed experimental dataset. All recordings use the same subset of 17 subjects and are spatio-temporally aligned across modalities.}
\begin{tabular}{c cc cc}
\toprule
 & \multicolumn{2}{c}{\textbf{Genuine}} & \multicolumn{2}{c}{\textbf{Replay}} \\
\cmidrule(lr){2-3} \cmidrule(lr){4-5}
 & Events & RGB & Events & RGB \\
\midrule

Device & EVK4 & FLIR & EVK4 & RealSense \\
FPS    & N/A  & 50   & N/A  & 30 \\
\#Subjects & 17 & 17 & 17 & 17 \\

\bottomrule
\end{tabular}
\label{tab:dataset_1}
\end{table}

\begin{table}[h]
\centering
\caption{Average duration and stimuli present in each type of clip}
\begin{tabular}{l cc cc}
\toprule
 & \multicolumn{2}{c}{\textbf{Genuine}} 
 & \multicolumn{2}{c}{\textbf{Replay}} \\
\cmidrule(lr){2-3} \cmidrule(lr){4-5}
Stimulus 
 & Duration(s) & \#Stimuli 
 & Duration(s) & \#Stimuli \\
\midrule

Fixation + Saccade 
 & 60 & 18 
 & 59.59 & 18 \\

Smooth Pursuit 
 & 100 & N/A 
 & 99.36 & N/A \\

\bottomrule
\end{tabular}
\label{tab:dataset_2}
\end{table}

\section{Evaluation Framework for Event-based Liveness Detection}

\subsection{Event-based data}

For each output clip, the event camera generates a set of events $E$. Each event $(x_k, y_k, t_k, p_k) \in E$ is a 4-tuple representing a logarithmic brightness change at pixel $(x_k, y_k)$ and time $t_k$ that exceeds a predefined threshold relative to the previous event at that pixel, where $p_k \in \{-1,1\}$ denotes the sign of the brightness change.

As part of an exploratory analysis of genuine and replay event-based clips, we compute several per-clip basic features that might characterize each set of videos. For each ROI (face, left eye, right eye) in a clip, we extract event-based statistics by discretizing the temporal stream into fixed-length windows of 33 ms. For each window, we compute the mean and standard deviation of the event rate and polarity balance. To account for the influence of the screen refresh rate and video playback temporal resolution in PAD scenarios, we introduce a per-pixel inter-event interval (IEI) measure. For each window, inter-event intervals are computed per pixel and summarized using the median. Spatial information within each window is aggregated via the median across pixels, and the temporal behaviour is characterized using the mean and standard deviation across windows.

\subsection{Event signal processing and frame representation} \label{stats}


Unlike the window-based descriptors mentioned above, which capture global statistical differences, we use the event activity profile~\cite{lenz2020} as a continuous estimate of event activity to analyse the fine temporal dynamics of eye movements:

\begin{equation}
A_{p}(t_i) =
A_{p}(t_{u_p}) \, e^{-\frac{t_i - t_{u_p}}{\tau}}
+ \frac{1}{\mu} \mathbb{I}[p_i = p]
\label{eq:activity}
\end{equation}
where $t_{u_p}$ denotes the timestamp of the most recent event
of polarity $p \in \{+1,-1\}$ occurring before $t_i$ and $\mathbb{I}$ is the indicator function. Setting $\mu = 1$, the proposed activity profile resembles an exponential moving average, extended to an event-driven formulation with continuous-time decay over asynchronous event arrivals. The activity is computed independently for each polarity channel $p \in \{+1,-1\}$ resulting in two signals $A_{+}$ and $A_{-}$.
Since event arrival times for each polarity are asynchronous and sparse,
we compute a piecewise linear interpolation of each activity series and
sample the resulting signal at discretized time steps $dt$ of 10 ms.  We also compute a polarity-agnostic activity profile $A(t)$ which aggregates all events independently of their polarity.

We also incorporate event-based frame representations to integrate spatio-temporal information, enabling the use of standard vision models. The Surface of Active Events (SAE)~\cite{7508476} encodes the recent temporal history of events within a spatial neighbourhood using exponential decay kernels, providing a spatial analogue of the temporal accumulation mechanism defined in Equation~\ref{eq:activity}:
\begin{equation}
S_p(x,y,t_i) =
\exp\!\left(-\frac{t_i - t_{x,y}^p}{\tau}\right)
\end{equation}
where $t_{x,y}^p$ is the timestamp of the most recent event at pixel $(x,y)$ with polarity $p$
\subsection{Evaluation methods} \label{sec:methods}

The considered models operate on event-based representations and differ in how temporal information is encoded and processed. To address saccade detection, Temporal Convolutional Networks (TCN)~\cite{lea2017tcn} are used to model temporal dependencies in event-activity time series. The proposed TCN architecture employs a deep stack of 1D dilated convolutions with residual connections along the temporal axis, enabling the aggregation of local temporal context while maintaining computational efficiency. The first stage includes two residual convolutional blocks that perform low-level feature extraction, increasing the input feature dimensionality to 32 and then 64 feature channels. Temporal context is then expanded using exponentially increasing dilation factors: the first stack of 4 residual blocks uses dilation rates from 2 to 16, and the second stack from 1 to 8. The final prediction layer is a 1D convolution followed by a sigmoid activation that outputs the probability of a saccade at each time step, preserving the temporal resolution of the input.

For face liveness detection, Spiking Convolutional Neural Networks (SCNN)~\cite{neftci2019scnn} are used to directly process event data. The event stream from a clip of spatial size $H \times W$ is divided into $T$ temporal bins and represented as a voxel grid, yielding a discretized event tensor of shape $[T,C,H,W]$, where polarity channels encode positive and negative events. The network consists of stacked convolutional blocks with leaky integrate-and-fire (LIF) neurons, enabling temporal dynamics through membrane potential integration. Feature maps are spatially aggregated and temporally averaged to produce the final prediction.
Vision Transformers (ViT-B16)~\cite{dosovitskiy2020vit} are also considered to incorporate temporal patterns from event-derived representations. Event streams are first converted into SAE frames, which are then processed by a transformer with a B16 backbone. This allows the model to capture global spatial dependencies while implicitly encoding temporal information through the input representation. 

\subsection{Evaluation Metrics}

We evaluate the proposed methods on two separate tasks: 
\begin{inparaenum}[a)]
    \item ocular movement detection (saccades and blinks), and
    \item face liveness detection.
\end{inparaenum}
For ocular movement detection, the problem is formulated as a temporal segmentation task. Following the evaluation protocol in~\cite{lea2017tcn}, predictions are matched to ground truth segments using temporal Intersection-over-Union (IoU), and a prediction is considered correct if its IoU exceeds the 0.5 threshold. Performance is reported using macro-averaged Precision, Recall, and F1-score computed over the matched segments.

Liveness detection is formulated as a binary classification problem (genuine vs. replay), and classification accuracy is reported as the primary metric. Additionally, biometric-specific error rates are considered, including the Attack Presentation Classification Error Rate (APCER), defined as the proportion of attacks incorrectly classified as bona fide, and the Bona Fide Presentation Classification Error Rate (BPCER), defined as the proportion of genuine samples incorrectly classified as attacks. We also report their average, the Average Classification Error Rate (ACER).

For each task, all models are trained and evaluated under the same protocol, enabling a direct comparison between different approaches. In both cases, we use the same split by subject with a ratio of 80:20 for the train and validation sets.

\subsection{Ground Truth annotations}

To validate the performance of saccades and blink detection, we had to manually annotate precise temporal segments for each genuine sequence in the RGBE-Gaze dataset. Since RGB and event cameras were synchronized, segmenting temporal blinks and saccades was possible by visually inspecting the RGB frames. We note that the resulting annotations include a margin of neutral activity surrounding each event, and therefore are not strictly bounded to the exact onset and offset times. A total of 173 blinks were annotated, considering those where more than half of the pupil was covered by the upper lid at the apex. 
Although each clip had an average of 18 stimuli displayed on the screen to elicit saccadic movements, we observed that many participants also performed additional saccades, possibly as part of the process of discovering the next stimulus. Those cases were deemed relevant for our study, while extremely short saccades of less than 40 ms (2 frames) were ignored. In total, 425 saccades were annotated.

It must be noted that the distribution of positive segments (for both saccades and blinks) is heavily unbalanced, reaching only a 3$\%$ from the total in the case of saccades. 

\begin{figure}[h]
    \centering
    \includegraphics[width=\linewidth]{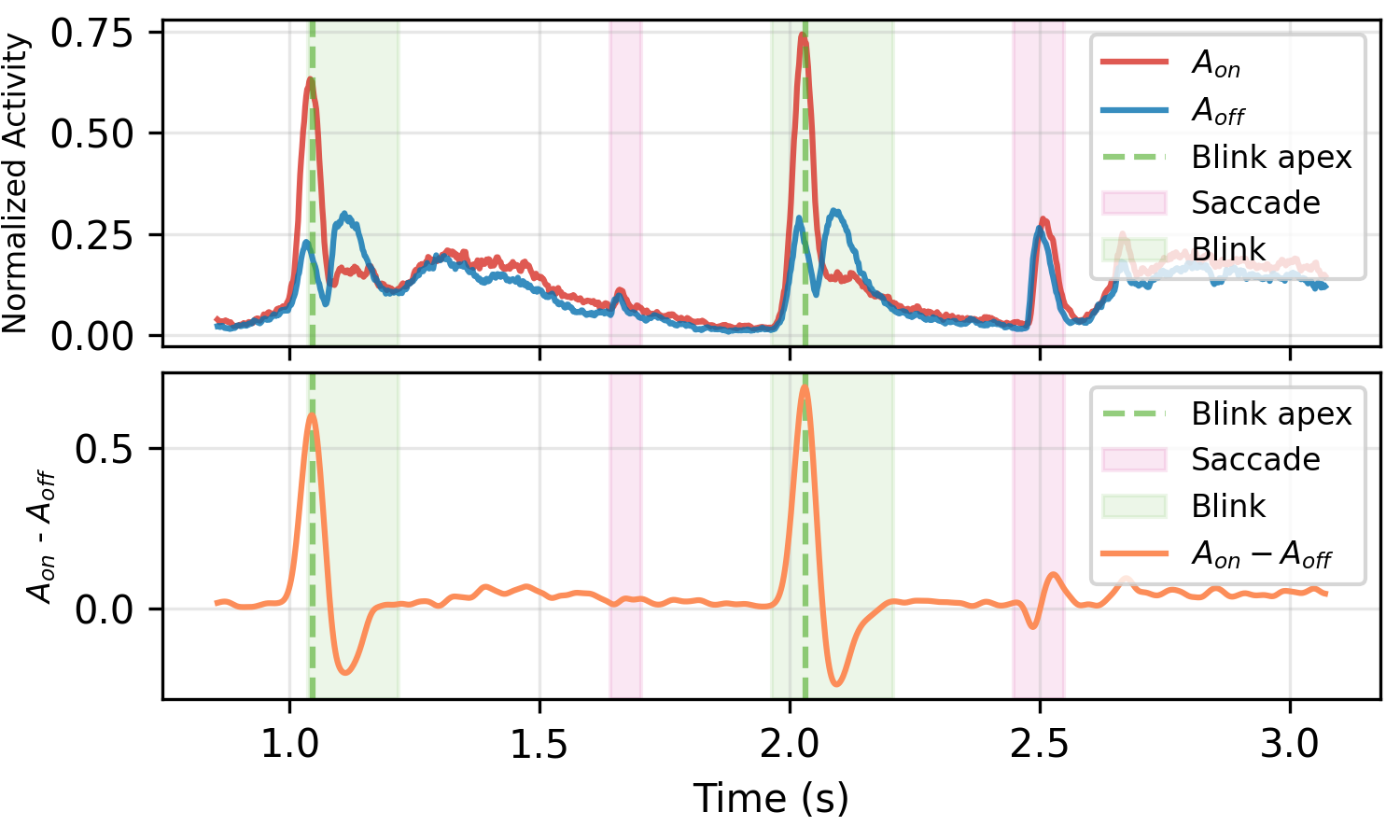}
    \caption{Normalized Activity profile for a saccade + fixation sample subsequence containing two blinks. Different from saccades, the temporal firm of blinks exhibits a distinctive double peak in both $A_{on}$ and $A_{off}$ activities, as well as a change in polarity balance. Zero-crossings in the smoothed signal $A_{on}$-$A_{off}$ represent the candidate timestamps for eyelid reopening.}
    \label{fig:event_blink_activity}
\end{figure}

\section{Results and Discussion} 
\label{sec:results}

\subsection{Ocular movements detection}
\begin{figure*}[h]
    \centering
    \includegraphics[width=\linewidth]{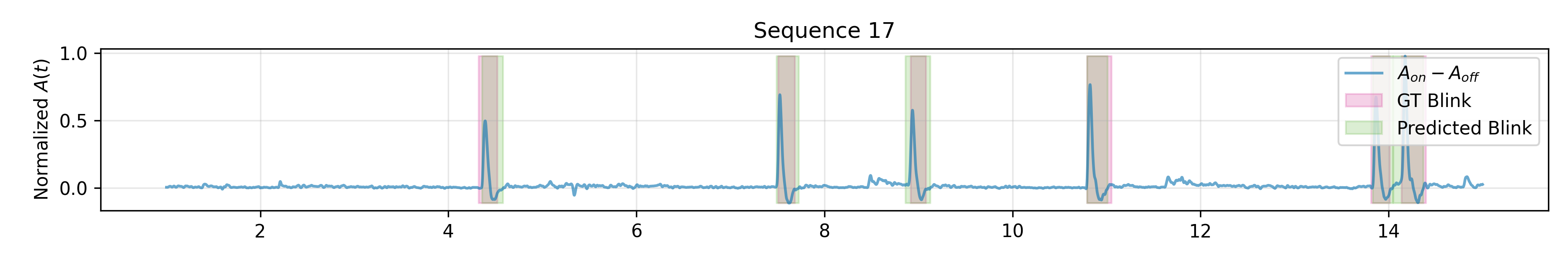}
    \includegraphics[width=\linewidth]{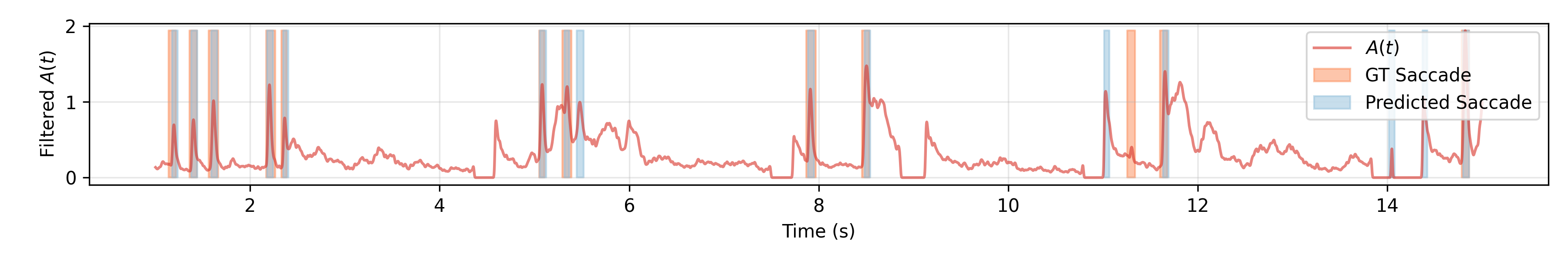}
    
    \caption{Normalized difference of positive and negative activity profile of a subsequence from subject 17 during the fixation + saccade experiment. The sequence contains multiple blinks detected by the blink detection algorithm (top). After blink suppression and normalization, saccades are segmented by identifying narrow, prominent peaks above a predefined threshold in the resulting $A(t)$ signal (bottom).}
    \label{fig:activity}
\end{figure*}

We analysed the viability of using blinks and saccades as possible ocular movements during the challenge-response stage. We framed this as a temporal segmentation problem, restricting the analysis only to genuine fixation + saccades clips. To this end, we analysed the effectiveness of both a simple signal processing approach and an end-to-end TCN model. In both cases, we represented the event streams using only the Activity Profile (see Equation~\ref{eq:activity}) computed over the left eye region, with $\tau = 10\,\text{ms}$ and a discretization step of $dt = 2\,\text{ms}$.

\subsubsection{Blink detection}

From the Activity profile, it emerged that the temporal firm of blinks follows a distinctive pattern as it was also noted by~\cite{lenz2020}\cite{Chen2021NeuroBiometricAE}. To illustrate the physical principle behind this, a fragment of a clip containing saccades and blinks is shown in Figure~\ref{fig:event_blink_activity}. 

In comparison with the upper eyelid, the interior of the eye is covered by more dark areas than brighter areas. This is mainly because of the camera position; the shadows from the eyelashes and the iris size compared to the sclera. When the eyelid closes, the positive contrast dominates the negative, producing a higher peak in the activity profile. After a short stationary period, the eyelid raises again, and the situation is reverted, but the peaks are shallower since reopening the eyelid takes longer. In the previous work of~\cite{lenz2020}, it was shown that this temporal signature of blinks could be exploited to detect faces at different scales. Here, we focused on the temporal segmentation of blinks given that faces were previously detected, and their scale is relatively stable.

To detect this temporal pattern, our approach consisted of searching for zero-crossings in the Gaussian smoothed $A_{on} - A_{off}$ signal. In this way, zeros become candidate timestamps indicating eyelid reopening. Then, we defined a temporal window around each candidate filtering those having one peak to the right and to the left with significant prominence. The sum of the peaks' widths is the estimation of the total duration. 

For the proposed method, we tuned the positive and negative peak prominence thresholds and set the window size to the 95th percentile of blink durations observed in the training set. We evaluated the detector on the validation set and achieved macro-averaged Precision and Recall of 97.62\% and 93.18\%, respectively, yielding an F1-score of 95.35\% at an IoU threshold of 0.5. From these results, we conclude that although the activity representation relies solely on the timing of events and ignores spatial information, the approach is suitable for blink detection in our remote-eye scenario with moderate head pose variations. Furthermore, we observed that some incomplete blinks also produce a reasonably strong activity, suggesting that this method may be suitable for applications in driver monitoring systems, such as drowsiness detection, among others.

\subsubsection{Saccade detection}

To establish a baseline method for temporal saccade segmentation, we proposed a peak detection algorithm based on the activity profile $A_t$. 

Because ocular activity is largely dominated by blinks, which can obscure the comparatively weaker saccadic signals, we first apply the previously described blink detection method as a preprocessing step, as can be seen in Figure~\ref{fig:activity}. Once blinking activity is removed, saccadic movements appear as relatively high spikes compared to the rest of the activity, presumably related to head movement. We then normalized $A_t$ and tuned the threshold to detect prominent peaks' width between 20 and 150 ms. 

\begin{table*}[h]
\centering
\caption{Performance comparison for event-based face liveness detection. APCER: Attack Presentation Classification Error Rate, BPCER: Bona Fide Presentation Classification Error Rate, ACER: Average Classification Error Rate.}
\label{tab:liveness_metrics}
\begin{tabular}{lccccccc}
\toprule
Method & Backbone & Data & Time Window & Top-1 Accuracy (\%) & APCER (\%) & BPCER (\%) & ACER (\%) \\
\midrule
SCNN & Conv. & Events (Voxel Grid) & Full & 95.37 & 4.20 & 5.10 & 4.65 \\ \hline\hline
SCNN & Conv. & Events (Voxel Grid) & 100 ms & 92.48 & 6.90 & 8.10 & 7.50 \\
ViT & B16 & Frames (RGB) & N/A & 92.15 & 8.20 & 7.60 & 7.90 \\
ViT & B16 & Events (SAE) & 100 ms & 90.62 & 10.20 & 8.50 & 9.35 \\
\bottomrule
\end{tabular}
\end{table*}

Alternatively, we followed an end-to-end approach based on the TCN architecture described in Section~\ref{sec:methods}. The model received a two-channel input consisting of the $A_{on}$ and $A_{off}$ signals of each clip without any preprocessing step other than standardization. For each signal, windows of 400 ms were extracted using a stride of 75 ms. The dilated convolution structure was chosen to ensure the effective receptive field encompasses the full input window. During evaluation, predictions from overlapping windows were averaged to produce a single probability estimate per time step. The model was trained for 100 epochs with a batch size of 64 and early stopping using the Adam optimizer with an initial learning rate of $10^{-3}$ and weight decay $10^{-4}$. The learning rate was scheduled using a ReduceLROnPlateau policy. To address class imbalance, the loss function incorporated a class-weighted focal loss, where the positive class weight was computed as the inverse class frequency. To improve evaluation, we optimized the decision threshold for predictions on the training set and removed predicted intervals shorter than 20 ms to reduce spurious detections. 

Results of these experiments are shown in the first row of Table~\ref{tab:saccades}. We found that, even after removing blinks, simple peak detection alone was not sufficient to achieve a strong F1-score, possibly due to additional noise caused by other transient head movements. However, further signal processing techniques would be worth exploring. In comparison, the TCN model outperformed this baseline, reaching an F1-score of 89.65\%.

\begin{table}[h]
\centering
\caption{Metrics for saccade and blink segmentation (IoU@0.5).}
\label{tab:ocular}
\begin{tabular}{llccc}
\hline
Task & Method & Precision & Recall & F1 \\
\hline
\multirow{3}{*}{Saccade Segmentation}
 & Peaks & 64.63 & 60.92 & 62.72 \\
 & TCN   & 89.65 & 89.65 & 89.65 \\
\hline
\multirow{1}{*}{Blink Segmentation}
 & Peaks & 97.62 & 93.18 & 95.35 \\
\hline
\end{tabular}
\label{tab:saccades}
\end{table}

\begin{figure}[h]
    \centering
    \includegraphics[width=\linewidth]{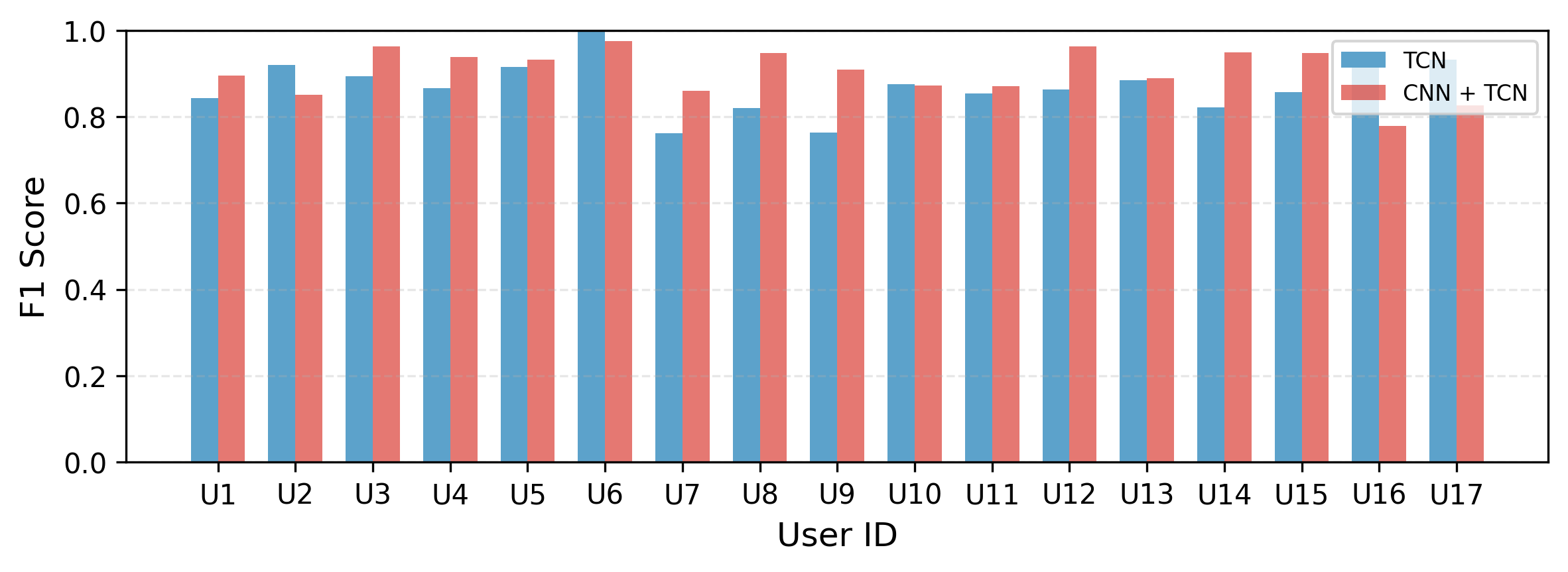}
   \caption{F1-score per subject under LOSO cross-validation. TCN uses activity profile while CNN+TCM uses SAE frame representations}
    \label{fig:final}
\end{figure}

Additionally, we evaluated generalization performance using leave-one-subject-out (LOSO) cross-validation. Hyper-parameters are fixed a priori based on previous experimentation and for each fold, the decision threshold is estimated using only the training subjects. We compared the previous TCN operating on the activity profile and a CNN + TCN model using SAE representations, which encode spatio-temporal event information as frame-like inputs. Results are reported in Figure~\ref{fig:final} as F1-score per subject, showing that incorporating spatio-temporal activity representations improves performance across subjects. The event activity based TCN achieves a macro F1-score of 86.92\% , while the CNN + TCN reaches 90.36\%.

In conclusion, we studied whether ocular movements—specifically blinks and saccades—could be reliably detected when used as liveness cues during the challenge-response phase of our system. Interestingly, a simple signal processing approach achieved reasonably good results for blink temporal segmentation. Saccade segmentation proved more challenging, but the TCN approach demonstrated the strongest performance.

\subsection{Face liveness detection}

We first analysed the distribution of the features described in Section~\ref{stats} to provide empirical evidence that temporal resampling and display-induced artifacts affect the perceived motion in replay attack clips.

\begin{figure}[h]
    \centering
    \includegraphics[width=\linewidth]{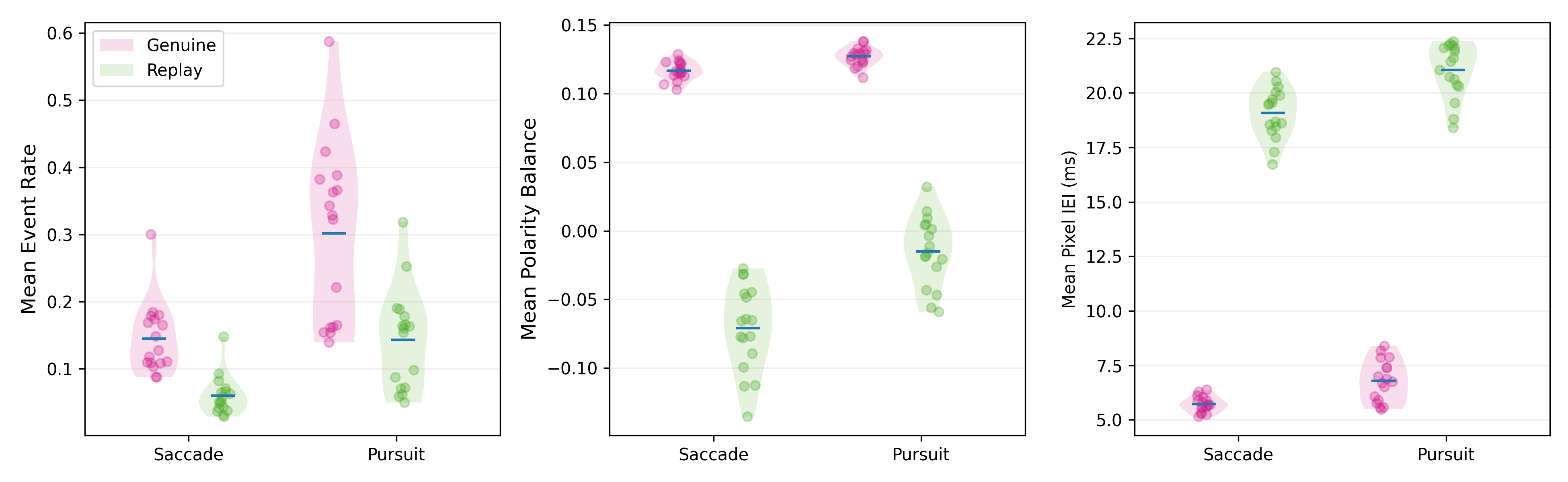}
    \caption{Each violin plot shows the distribution of Mean Event Rate, Mean Polarity Balance, and Mean Pixel IEI in genuine and replay attack recordings. Features are obtained from the face bounding box region. These statistical features reinforce the presence of temporal discretization effects and illumination artifacts introduced by the screen.}
    \label{fig:stats}
\end{figure}

As shown in Figure~\ref{fig:stats}, some of these features appear to reasonably differentiate genuine recordings from replay videos and could therefore support the training of a simple classifier. We observe that the mean event rate is typically lower in replay attacks due to the limited brightness of the display compared to genuine recordings. In addition, the mean pixel IEI indicates that intervals between events tend to lie close to the expected duration of each frame in a 50 fps video. In contrast, genuine recordings are not constrained by the frame rate and therefore exhibit much lower IEI values.
However, we argue that they might be insufficient to build a robust detection system. For example, an attacker could ramp up the brightness of the screen to match the desired event rate; inject additional noise to the RGB video to induce fake events; use a different screen technology and temporal upsample the RGB video, allowing them to better match the genuine Pixel IEI distribution.
The results of training the SNN and ViT models for this task are presented in Table~\ref{tab:liveness_metrics}. The convolutional SNN trained with the largest temporal window achieved the best performance across all evaluation metrics, reaching a top-1 accuracy of 95.37\% and an ACER of 4.65\%. Notably, in contrast to the RGB modality, this model outperformed the ViT model trained on frames yielding a promising direction for event-based approaches for liveness detection.

\section{CONCLUSIONS AND FUTURE WORKS}

In this work, we introduced an event-based approach for face liveness detection based on temporal ocular dynamics. The proposed framework, evaluated on a replay attack dataset derived from RGBE-Gaze, showed that different event-based representations enable reliable separation between genuine and replayed sequences, achieving up to 95.37\% top-1 accuracy with a Spiking Convolutional Neural Network.

Beyond liveness detection, we also showed that temporal modelling of event signals allows accurate temporal segmentation of ocular movements, supporting the feasibility of using saccades and blinks as part of a challenge-response mechanism. These findings highlight the potential of event-based sensing for low-latency and privacy-preserving biometric systems.

Despite these promising results, several directions remain open. First, the current study focuses on replay attacks using a single display configuration. Future work will extend the dataset to include a broader range of presentation attack instruments, such as different screen technologies, refresh rates, resolutions, and illumination conditions. Additionally, incorporating other attack types, such as print attacks, mask-based attacks, IA-generated forgeries, and more advanced display systems, would allow a more comprehensive evaluation.

Second, the current dataset is limited in scale and diversity. Expanding it to include more subjects, environmental variability, and acquisition setups will be essential to assess generalization. In particular, inter-dataset evaluation remains an open problem due to the scarcity of event-based facial datasets, and future efforts should aim at enabling cross-dataset benchmarking.

Finally, future work will explore fully event-based pipelines by removing the dependency on RGB for face registration, as well as investigating more advanced temporal models for jointly capturing ocular dynamics and liveness cues. Overall, this work provides initial evidence that temporal ocular dynamics in event data constitute a promising direction for robust liveness detection, motivating further research in this area.

\section*{Acknowledgment}

The authors acknowledge the use of AI-based tools, such as Grammarly, for assistance in editing, grammar enhancement, and spelling checks during the preparation of this manuscript.


{\small
\bibliographystyle{ieee}
\bibliography{egbib}
}


\end{document}